\documentclass[runningheads]{llncs}

\usepackage[T1]{fontenc}
\usepackage{CJK} % Chinese
\usepackage[T1]{fontenc}
\usepackage{graphicx} % to reize tabel and figure
\usepackage{color}
\usepackage{booktabs}
\usepackage{array}
\usepackage{latexsym}
\usepackage{multirow}
\usepackage{enumitem}
\usepackage{hyperref}

\usepackage{pifont} % check mark and cross

\usepackage{acronym}
\acrodef{NLP}{natural language processing}
\acrodef{MPMCP}{multi-person multi-charge prediction}
\acrodef{LJP}{legal judgment prediction}
\acrodef{LLM}{large language model}
\acrodef{IQR}{interquartile range}

\newcommand{\cmark}{\ding{51}}%
\newcommand{\xmark}{\ding{55}}%

\newcommand{\todo}[1]{\textcolor{black}{#1}}
\newcommand{\jp}[1]{\textcolor{black}{#1}}

\begin{document}
\title{MultiJustice: A Chinese Dataset for Multi-Party, Multi-Charge Legal Prediction}
% \author{Anonymous authors}
% \input{Content/authors}
%
\titlerunning{MultiJustice}

\author{
Xiao Wang\inst{1}\thanks{Equal contribution}  \and
Jiahuan Pei\inst{2}$^\star$ \and
Diancheng Shui \inst{3} \and
Zhiguang Han \inst{4} \and
Xin Sun \inst{5} \and
Dawei Zhu  \inst{1} \and
Xiaoyu Shen \inst{6}\thanks{Corresponding author} 
}

\authorrunning{X. Wang and J. Pei et al.}

\institute{
Saarland Informatics Campus, Saarland University, Saarland, Germany
\email{xiaowang@lst.uni-saarland.de} \and
Vrije Universiteit Amsterdam, Amsterdam, Netherlands
\email{j.pei2@vu.nl} \and
Wuhan University, Wuhan, China \and
Nanyang Technical University, Singapore, Singapore \and
University of Amsterdam, Amsterdam, Netherlands \and
Eastern Institute of Technology, Ningbo, China 
\email{xyshen@eitech.edu.cn}
}


\maketitle              % typeset the header of the contribution

\begin{abstract}
\jp{
\Ac{LJP} offers a compelling method to aid legal practitioners and researchers. 
However, the research question remains relatively underexplored: 
\textit{Should multiple defendants and charges be treated separately in \ac{LJP}?}
To address this, we introduce a new dataset, namely \ac{MPMCP}, and seek the answer by evaluating the performance of several prevailing legal \acp{LLM} on four practical legal judgment scenarios:
(S1) single defendant with a single charge,
(S2) single defendant with multiple charges,
(S3) multiple defendants with a single charge, and
(S4) multiple defendants with multiple charges.
We evaluate the dataset across two \ac{LJP} tasks, i.e., charge prediction and penalty term prediction.
We have conducted extensive experiments and found that the scenario involving multiple defendants and multiple charges (S4) poses the greatest challenges, followed by \todo{S2, S3, and S1}. 
The impact varies significantly depending on the model. 
For example, in S4 compared to S1, InternLM2 achieves approximately \todo{4.5\%} lower F1-score and \todo{2.8\%} higher LogD, while Lawformer demonstrates around \todo{19.7\%} lower F1-score and \todo{19.0\%} higher LogD.
Our dataset and code are available at \url{https://github.com/lololo-xiao/MultiJustice-MPMCP}.
}

%%%% textual info for submission
\if0
Legal judgment prediction offers a compelling method to aid legal practitioners and researchers. 
However, the research question remains relatively under-explored: Should multiple defendants and charges be treated separately in LJP? 
To address this, we introduce a new dataset namely multi-person multi-charge prediction (MPMCP), and seek the answer by evaluating the performance of several prevailing legal large language models (LLMs) on four practical legal judgment scenarios:
(S1) single defendant with a single charge,
(S2) single defendant with multiple charges,
(S3) multiple defendants with a single charge, and
(S4) multiple defendants with multiple charges.
We evaluate the dataset across two LJP tasks, i.e., charge prediction and penalty term prediction.
We have conducted extensive experiments and found that the scenario involving multiple defendants and multiple charges (S4) poses the greatest challenges, followed by S2, S3, and S1. 
The impact varies significantly depending on the model. 
For example, in S4 compared to S1, InternLM2 achieves approximately 4.5\% lower F1-score and 2.8\% higher LogD, while Lawformer demonstrates around 19.7\% lower F1-score and 19.0\% higher LogD. Our dataset and code are available at https://github.com/lololo-xiao/MultiJustice-MPMCP.
\fi

\end{abstract}

\section{Introduction}
\jp{
The emergence of \acp{LLM} has significantly accelerated research across a range of critical and specialized domains, including 
psychotherapy~\cite{sun2024eliciting,sun2025rethinking,sun2024script},
medicine~\cite{yan2022remedi,wang2023pre}, and 
industrial coaching~\cite{pei2024autonomous}. In the legal field, \acp{LLM} have similarly emerged as a prevailing paradigm. 
For example, 
DISC-LawLLM~\cite{yue2023disclawllm} excels in providing comprehensive legal consultation, and LawBench~\cite{fei2023lawbench} attracts an increasing number of \acp{LLM} for evaluation of legal tasks.
\textit{\Acf{LJP}} is a crucial task for intelligent legal assistants, which aims to predict case outcomes based on factual descriptions~\cite{cui2022survey}. 
These outcomes typically encompass the types of charges and terms of penalty in the study of China's criminal law. 
}

\jp{
However, complex judgment prediction involving multiple defendants and multiple charges is common but highly challenging in real-world scenarios:
In TOPJUDGE~\cite{zhong2018legal}, these complex cases are fully neglected to explore relationales between various subtasks.
In MAMD~\cite{pan2019charge}, there are approximately 30.32\% of cases involve multiple defendants.
In MultiLJP~\cite{lyu2023multi}, 89.58 \% of the cases the defendants received different judgments for at least one of the subtasks in the multi-defendant \ac{LJP} task.
}

\begin{figure}[t!]
  \includegraphics[trim={0 0 0 0},clip,width=\linewidth]{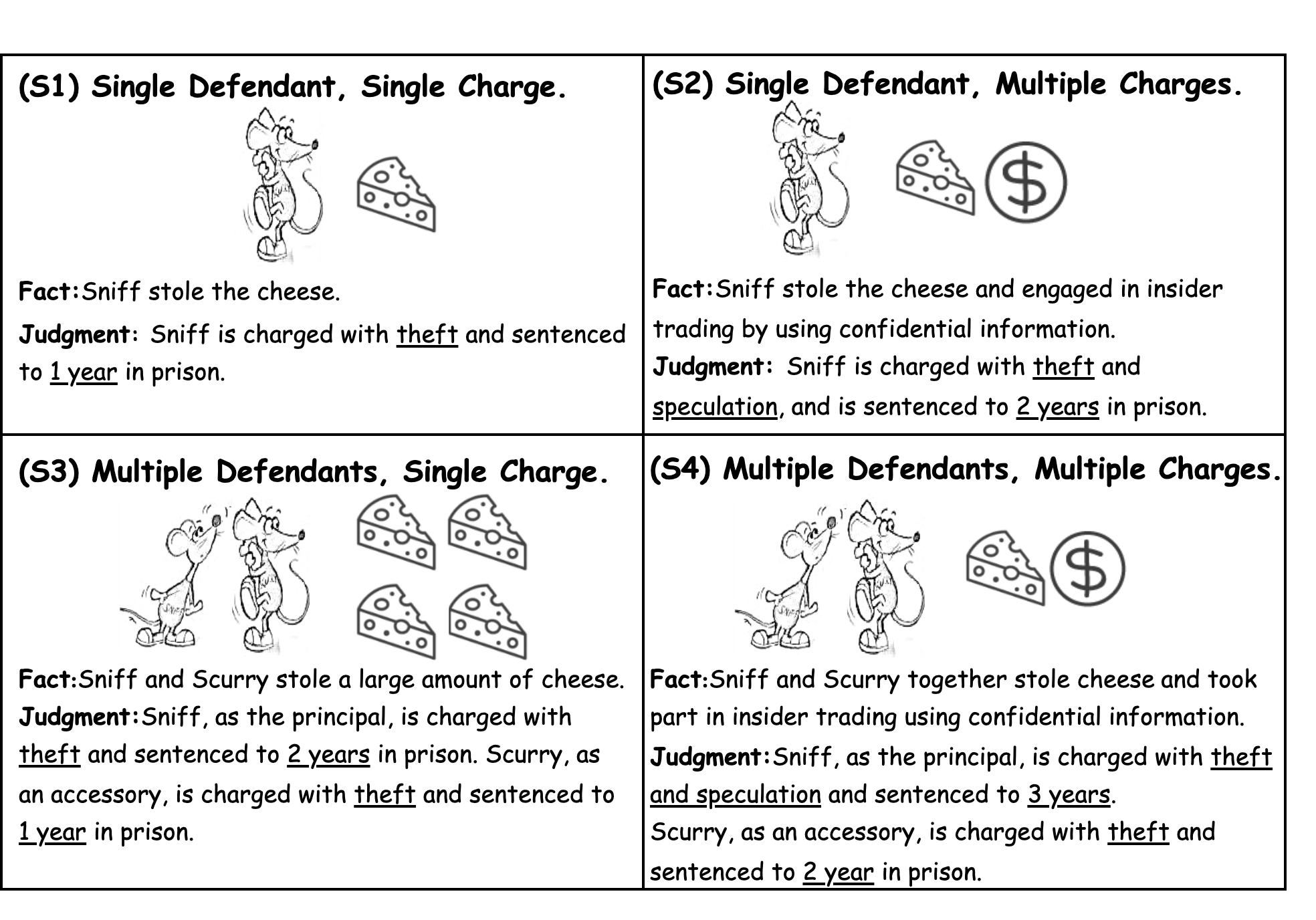}
  \caption{An illustration of the various charges and terms of penalty in four practical legal judgment scenarios: 
  (S1) single defendant with a single charge, 
  (S2) single defendant with multiple charges, 
  (S3) multiple defendants with a single charge, and 
  (S4) multiple defendants with multiple charges.
  }
  \label{fig:motivation}
\end{figure}

\jp{To address this gap, we introduce \ac{MPMCP} dataset with four practical scenarios, as illustrated in Figure~\ref{fig:motivation}.
For example, in (S4), the two defendants (i.e., Sniff and Scurry) should receive different outcomes (i.e., charges and penalty terms) based on the description of a fact involving two charges (i.e., theft and speculation).
Unlike (S1), the factual description in (S4) involves more defendants and charges and provides more details, such as activities (i.e., stealing cheese and insider trading) and methods (i.e., using confidential information).
As the number of defendants and charges increases, the complexity of the factual description also escalates, presenting greater challenges for prediction models.  
With an exploratory study of the proposed dataset, we seek to answer the main research question: 
\begin{quote}
\textit{How well can legal judgment prediction (LJP) models—especially large language models—generalize across realistic and increasingly complex legal case scenarios involving multiple defendants and charges?}
\end{quote}
}

\jp{
We use five prevailing open-source \acp{LLM} (i.e., MT5, MBERT, RoBERTa, LawFormer, and InternLM2) as benchmark models for generating charges and penalty terms across four scenarios in Chinese \ac{LJP}.
We also analyze the performance of InternLM2 variants under multiple settings 
% (\todo{i.e., fine-tuning, multi-task, zero-shot, one-shot}) 
to provide empirical insights into how these settings influence different scenarios.
The main findings are that scenarios involving multiple defendants and multiple charges (S4) pose the greatest challenges, followed by S2, S3, and S1; The overall performance drops dramatically as the complexity of the scenario increases, although the relative impact varies significantly depending on the model.
Our contributions include:
(1) MPMCP dataset, which encompasses four practical legal judgment scenarios involving multiple defendants and multiple charges.
(2) An exploratory study on benchmark models and the variant settings in different scenarios.
}

\section{Related Work}
\Acf{LJP} is a critical task for smart legal assistants, which aims to predict the outcomes of legal cases given the description of facts~\cite{cui2022survey}.
These outcomes usually include the types of the charge(s) and terms of penalty.
Due to differences in national legal systems~\cite{sznycer2020origins}. 
Specifically, our study specifically focuses on criminal legal cases within the Chinese judicial framework.

Most existing research has introduced datasets and models to advance \ac{LJP}, as summarized in Table~\ref{tab:datasets}.
\begin{table*}[htb!]
\centering
\setlength{\tabcolsep}{5pt}
\footnotesize
\resizebox{\linewidth}{!}{
\begin{tabular}{@{}lccccrrrr@{}}
\toprule
\multirow{2}{*}{Dataset} 
& \multicolumn{2}{c}{Defendant} & \multicolumn{2}{c}{Charge} 
& \multirow{2}{*}{\#Case} & \multirow{2}{*}{\#Charge} & \multirow{2}{*}{\#Term} & \multirow{2}{*}{\#Article} \\ \cmidrule(lr){2-3} \cmidrule(lr){4-5}
& Single & \multicolumn{1}{c}{Multiple} 
& Single & \multicolumn{1}{c}{Multiple} & & & \\
\midrule
CAIL2018~\cite{xiao2018cail2018} & \cmark & \xmark & \cmark & \xmark & 2,676,075 & 202 & 3 & 183 \\
TOPJUDGE-CAIL~\cite{zhong2018legal} & \cmark & \xmark & \cmark & \xmark &  113,536 &  99 & 3 & 98 \\
MAMD~\cite{pan2019charge} & \cmark & \cmark & \cmark & NA & 164,997 & NA & NA & NA \\
% \midrule
CAIL-Long~\cite{xiao2021lawformer} & \cmark & \xmark & \cmark & \xmark & 229,505 & 201 & 5 & 244 \\
RLJP~\cite{wu2022towards} & \cmark & \xmark & \cmark & \xmark & 89,768 &  48 & 1 & 95 \\
SLJA-COR~\cite{deng2023syllogistic}  & \cmark & \xmark & \cmark & NA & 11,239 & 80 & 5 & 124  \\ 
MultiLJP~\cite{lyu2023multi}  & \cmark & \cmark & \cmark & NA & 23,717 & 23 & 11 & 22 \\ % Total defendants 80,477
\midrule
MPMCP (Ours)  & \cmark & \cmark & \cmark & \cmark & 20,000 & 306 & 1 & 234 \\
\bottomrule
\end{tabular}
}
\caption{Comparable public datasets for legal judgment prediction involving single vs. multiple defendants and charges.
The symbol ``\cmark'' indicates that a characteristic is explicitly covered in a dataset, ``\xmark'' indicates that it is explicitly not covered, and ``NA'' denotes ``not applicable'' as it is not explicitly concerned in the reference work.
}
\label{tab:datasets}
\end{table*}

% TopJudge~\cite{zhong2018legal}
% In real-world scenarios, there are some cases
% with multiple defendants and multiple charges,
% which will increase the complexity of judgment
% prediction. As our model aims to explore the effectiveness of considering topological dependencies between various subtasks, we filter out these
% cases and leave them as our future work
%%%
% CAIL2018~\cite{xiao2018cail2018} release a large-scale legal dataset for fundamental 
% \ac{LJP} research considering a single defendant with a single charge.
% They implement several conventional text classification models (i.e., TFIDF+SVM, FastText, CNN) to facilitate the development and benchmarking of \ac{LJP} models.
The CAIL2018 dataset~\cite{xiao2018cail2018} is one of the foundational benchmarks in this field. It contains large-scale criminal cases, each involving a single defendant and a single charge. Alongside the dataset, the authors evaluated standard text classification baselines (e.g., TF-IDF + SVM, FastText, and CNN), laying the groundwork for subsequent research.
%%%
% TOPJUDGE-CAIL~\cite{zhong2018legal} highlight the challenge of complex judgment prediction involving multiple defendants and multiple charges in real-world scenarios. 
% However, their study focuses on exploring topological dependencies between various subtasks, without handling these complex cases.
TOPJUDGE-CAIL~\cite{zhong2018legal} the complexity of real-world legal scenarios involving multiple defendants and charges. While they explore topological dependencies across subtasks such as charge prediction and sentencing, their dataset and methods do not directly model multi-defendant or multi-charge situations.
%%%
% \cite{pan2019charge, lyu2023multi} focus on multi-defendant legal judgment prediction, without distinguishing whether the charges are single or multiple.
MAMD~\cite{pan2019charge} and MultiLJP~\cite{lyu2023multi} extended the task to multi-defendant cases. However, their works do not explicitly distinguish between single and multiple charges per defendant, limiting their applicability in more nuanced legal contexts.
%%%
CAIL-Long~\cite{xiao2021lawformer} introduces Lawformer, a pre-trained language model specifically designed for Chinese legal long documents. 
% This model addresses the challenges associated with processing lengthy legal texts, improving the accuracy of judgment predictions by leveraging a hierarchical transformer architecture.
By leveraging a hierarchical Transformer architecture, it improves performance on tasks involving lengthy factual descriptions.
%%%
% RLJP~\cite{wu2022towards} generate rationales and outcomes separately to enhance the interactivity and interpretability of legal judgment. 
RLJP~\cite{wu2022towards} advanced the interpretability of \ac{LJP} by separately generating legal rationales and outcomes, enhancing transparency and user trust in AI-based judgment systems.
%%%
% MultiLJP~\cite{lyu2023multi} focuses on multi-defendant legal judgment prediction via hierarchical reasoning. 
% This work highlights the complexity of cases involving multiple defendants, proposing methods to handle such scenarios effectively by capturing the interactions between different defendants within a case.
% MAMD~\cite{pan2019charge} deals with charge prediction for multi-defendant cases using a multi-scale attention mechanism. 
% This approach enhances the model's ability to distinguish between various charges in complex cases, providing more accurate predictions for each defendant.
SLJA~\cite{deng2023syllogistic} present a method for syllogistic reasoning in legal judgment analysis and provide several \acp{LLM} as benchmarks.
% SLJA~\cite{deng2023syllogistic} proposed a syllogistic reasoning framework that mimics human legal logic, offering a benchmark for evaluating large language models (\acp{LLM}) in legal reasoning tasks.
This approach emphasizes logical reasoning by structuring legal arguments in a way that mimics human legal reasoning, thereby improving the interpretability and accuracy of LJP models.
More recently, MUD~\cite{wei2024through} focus on interpretable charge prediction and reasoning with fine-grained annotations via a symbolic-neural model on a new multi-defendant dataset.
Instead, our work offers a broader scenario-driven evaluation framework that benchmarks generalization and robustness across multiple real-world legal complexities using LLMs.
% More recently, \citet{li2024multiscale} proposed a multiscale graph reasoning network for capturing both case-level and defendant-level dependencies in multi-party legal prediction, significantly improving prediction accuracy and coherence.
% \citet{zhang2024lawgraph} introduced LawGraph, a structured legal knowledge graph that augments LJP models with relational legal knowledge, facilitating better generalization across complex case structures.
% \citet{hou2024llmjudge} explored few-shot LJP with large language models, demonstrating promising results on legal reasoning with limited training data, though lacking robust handling of multi-defendant, multi-charge scenarios.

% To sum up, these works address challenges such as handling long documents, and multi-defendant cases and enhancing logical reasoning with rationales.
% However, none of those works can fairly compare the difference between the four practical scenarios proposed in this study.

To sum up, prior studies have tackled challenges such as processing long legal documents, improving logical reasoning, and modeling multi-defendant cases. However, none of these works systematically address the distinctions between the four practical scenarios defined in our study—specifically, combinations of single/multiple defendants and single/multiple charges. Our work aims to fill this gap by proposing a new dataset and evaluation framework that fairly compares performance across these real-world complexities.

\section{Dataset Construction}
\subsection{Raw data collection} %% How did you collect the data
We constructed the MPMCP dataset using first-instance documents collected from China Judgments Online\footnote{\url{https://wenshu.court.gov.cn/}}, covering the period from 1998 to 2021.
\jp{We exclusively obtain criminal cases with judgment outcomes and retain documents that clearly identify defendants, provide factual descriptions, and include charges, penalty terms, and applicable legal articles.}
% We remain exclusively criminal cases with judgment results, filtering the other types.
% For the LJP task, we focused exclusively on criminal cases with judgment results. 
% Therefore, we retained only the relevant documents that clearly identified the defendants, provided a fact description, and included legal judgment outcomes such as the charges, the term of the penalty, and the applicable legal articles.

\subsection{Data Extraction} %%[Name] (related to main procedure)
\jp{
We utilize regular expressions to directly extract relevant facts, applicable legal articles, charges, and penalty terms from four sections in a document, identified by inherent keyphrases, e.g.,
``Upon trial, it was found'', ``This court believes'', and ``The judgment is as follows''.
}
 % Each original legal document is well-structured and can be divided into distinct parts based on its inherent keywords. Therefore, once we have obtained the raw data, we can directly use regular expressions to extract relevant facts, applicable legal articles, charges, and prison terms from the corresponding sections of the document.  
 % Specifically, each standard legal document contains three essential keywords: \textit{``Upon trial, it was found''}, \textit{``This court believes''}, and \textit{``The judgment is as follows''}. 
 % These keywords collectively divide the document into four main sections. 
 The first section provides a basic introduction to the case, which we do not consider relevant for dataset construction.
 The second section summarizes the facts of the case as determined by the court, based on statements from the parties involved, evidence presented, and court inquiries. 
 This section is typically used as input for the LJP models. 
 The third section contains the judge's explanation of the applicability of the law, including the legal articles referenced throughout the judgment process. 
 The final section details the judgments for each defendant, including the charges and the corresponding prison terms. 
 % Each contains five fields: defendant, criminal facts, applicable legal articles, charges, and penalty terms. 
 % Similar to previous methods for constructing datasets, we used regular expressions to directly extract the necessary information from the relevant sections.  
\jp{
 Notably, we preserve all defendants and their corresponding judgments for each case to ensure the dataset accurately reflects the actual conditions of judicial rulings. 
 }

\jp{
To ensure data quality, we mask any content within the extracted factual texts that precisely matched the names of the charges to prevent information leakage. 
We randomly select 5,000 cases for each of the 4 scenarios and manually assess approximately 5\% of the data to ensure the inclusion of 20,000 qualified cases in the final dataset.
}

 % Unlike previous approaches that only retained one-defendant cases, we preserved all defendants and their corresponding judgments for each case to ensure the dataset more accurately reflects the actual conditions of judicial rulings.
 % For our proposed dataset, we further divided it into four different settings based on the number of defendants and charges involved in each case. Specifically, these settings include: (1) one defendant with one charge, (2) one defendant with multiple charges, (3) multiple defendants with one charge, and (4) multiple defendants with multiple charges. 
 % Due to the presence of multiple defendants and multiple charges, MPMCP presents more complex scenarios, significantly increasing the difficulty of the LJP task.  
 
 % Additionally, to further ensure the quality of the dataset, we masked any content within the extracted fact texts that might exactly match the names of the charges to prevent information leakage. 
 % Furthermore, the top 102 legal articles in Chinese criminal law are not related to specific charges. Therefore, we excluded these legal articles and charges from the dataset. 
 
 % Finally, for each setting, we randomly selected 5,000 qualifying cases to form the final dataset, which will be used for training and evaluating the LJP models.

\subsection{Data statistics} %
\begin{figure*}[htb!]
    \centering
    \includegraphics[width=1.02\linewidth]{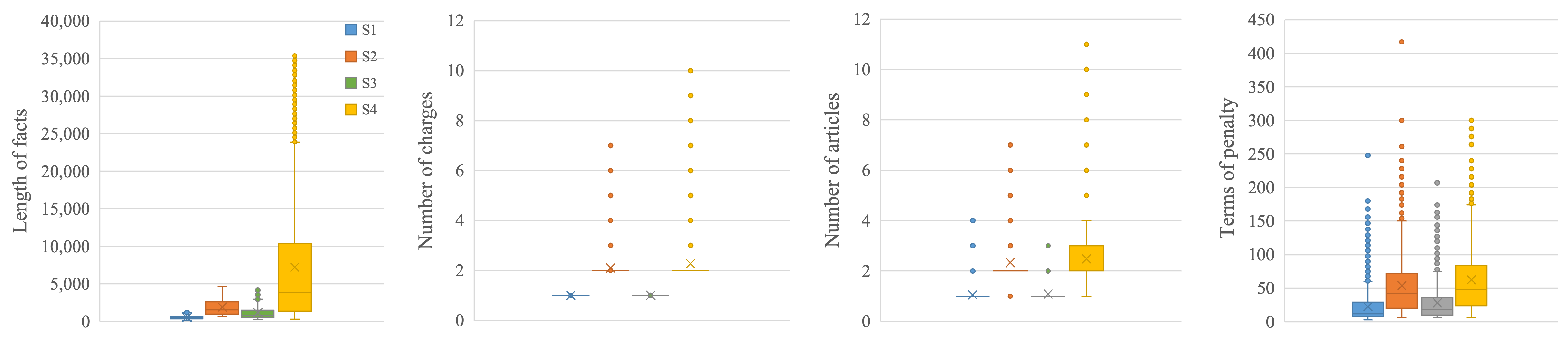}
    \caption{Box plots over the MPMCP dataset depict variations across four scenarios (S1, S2, S3, S4) for (a) number of facts, (b) number of charges, (c) number of legal articles, and (d) terms of penalty.
    In each box plot, the median is denoted by a line, and the mean value is marked by an ``$\times$''.
    }
    \label{fig:statistics}
\end{figure*}

Figure~\ref{fig:statistics} depicts the statistics of the proposed dataset.
We observe that:
First, the length of facts exhibits significantly higher median and mean values in (S4) compared to (S1, S2, S3), with the largest \acf{IQR} indicating diverse lengths. 
Similarly, this trend is observed in ``terms of penalty'' and ``number of articles'', where (S4) exhibits greater variability and higher median, mean, \acs{IQR} values compared to (S1, S2, S3).
This suggests that in (S4), the legal cases are more complex.
Second, the number of charges is predominantly concentrated on 1-2 charges. 
Compared to (S1, S3) involving only 1 charge per case, scenarios (S2, S4) exhibit an average of 2 charges per case, with several outliers ranging from 3 to 10 charges.
See Table~\ref{tab:data_samples} for the examples of our dataset in four different settings.

\begingroup
\begin{table*}[htb!]
    \centering
    \setlength{\tabcolsep}{5pt}
    \scriptsize
    \begin{tabular}{p{\linewidth}}
        \toprule
            \vspace{-2mm}
        \textbf{S1 (Single Defendant Single Charge)}\\
        \midrule
        \textbf{Defendant}: A1 \\
        \textbf{Fact}: Between October 7 and 20, 2019, \textit{\textbf{defendant A1 stole}} a total of 2,140 yuan from victims A2, A3, and A4 in Hanjiang District, Putian City. A was arrested on October 22, and the stolen cash was recovered and returned. On March 16, 2020, A signed a plea agreement. During the trial, A did not dispute the facts. The evidence was sufficient to confirm A's crimes.\\
        \textbf{Legal Judgment}:\\
        \ding{172}\textbf{Charges:}Theft \\
        \ding{173}\textbf{Penalty Term:} 8 Months \\
        \midrule
        \vspace{-1mm}
        \textbf{S2 (Single Defendant Multiple Charges)} \\
        \midrule
        \textbf{Defendant}: B1 \\
        \textbf{Fact}: On November ..., \textit{\textbf{B1 sold drugs }}to B2 near a hospital in....B1 was caught with 200 yuan and one packet of heroin... B1 did not contest the facts; the evidence was sufficient.

        On March ..., \textit{\textbf{B1 injured}} B4 during a dispute, \textit{\textbf{causing minor injuries}}... B1 did not contest the facts; the evidence was sufficient.\\
        \textbf{Legal Judgment}:\\
        \ding{172}\textbf{Charges:}Trafficking Drugs, Intentional Injury \\
        \ding{173}\textbf{Penalty Term:} 10 Years \\
        \midrule
        \vspace{-1mm}
        \textbf{S3 (Multiple Defendants Single Charge)} \\
        \midrule
        \textbf{Defendant}: C1 \\
        \textbf{Fact}:On May 20, 2019, \textit{\textbf{C1 and C2}} conspired to \textit{\textbf{steal}} at MingmenShijia Community... \textit{\textbf{C1 stole}} 100 yuan and a gold pendant from C3's home... Items were not recovered. \textit{\textbf{C1 and C2}} confessed; evidence was sufficient.\\
        \textbf{Legal Judgment}:\\
        \ding{172}\textbf{Charges:}Theft \\
        \ding{173}\textbf{Penalty Term:} 11 Months \\
        \midrule
        \vspace{-1mm}
        \textbf{S4 (Multiple Defendants Multiple Charges)} \\
        \midrule
        \textbf{Defendant}: D1  \\
        \textbf{Fact}:Between April and August, \textit{\textbf{D1 and D2 placed 27 gambling machines}} in Taizhou, earning 68,323 yuan. \textit{\textbf{D1 earned 20,000 yuan}}, D2 ... D1 also placed one machine alone, paying a 2,100 yuan bribe.
        
        In August, \textit{\textbf{D3 contacted D1 and D2 to sell over 40 gambling machines}} ..., earning 180,000 yuan...The evidence was sufficient, and all three had no objections.\\
        \textbf{Legal Judgment}:\\
        \ding{172}\textbf{Charges:}Operating a Gambling Den, Illegal Business Operations \\
        \ding{173}\textbf{Penalty Term:} 4 Years and 3 Months \\
        \bottomrule
    \end{tabular}
        \caption{
  Examples of data in four scenarios of MPMCP dataset.
    }
    \label{tab:data_samples}
\end{table*}
\endgroup

\section{Experimental Setup}
% In this section, we carry out the implementation and assessment of various baselines across two subtasks of LJP, encompassing charges, and prison terms

\subsection{Benchmark Models}
We leverage the following five prevailing open-source \acp{LLM} for Chinese  \ac{LJP} as benchmark models to generate outputs in four scenarios.
% that demonstrate performance on our dataset.

\noindent \textbf{MT5}~\cite{xue2021mt5}, a T5 variant with multilingual capabilities, pre-trained on a novel dataset derived from Common Crawl, encompassing 101 languages.

\noindent \textbf{MBERT}~\cite{devlin2019bert}, a BERT model pre-trained on 104 of the most resource-rich languages in Wikipedia, supporting multilingual functionality.

\noindent \textbf{RoBERTa
% (Robustly optimized BERT pretraining Approach)
}~\cite{liu2019roberta},
a variant of the BERT~\cite{kenton2019bert} with modifications to training dynamics.
% a model trained with dynamic masking, FULL-SENTENCES without NSP loss, large mini-batches, and a larger byte-level BPE, aiming to improve BERT implementation.

\noindent \textbf{Lawformer}~\cite{xiao2021lawformer}, a longfomer-based model pre-trained using extensive Chinese legal long case documents on a large scale

% \noindent \textbf{InternLM2}~\citep{cai2024internlm2}, a model build upon internlm2-base, further pre-trained on specific domain corpora. 
% It has demonstrated outstanding performance in evaluations within its designated field while maintaining excellent general language capabilities. 

\noindent \textbf{InternLM2}~\cite{cai2024internlm2}, built upon internlm2-base and additionally pre-trained on domain-specific corpora, excels in its designated field evaluations while retaining strong general language capabilities.

\subsection{Evaluation Metrics}

\jp{
We evaluate the generated legal judgment results in terms of charge prediction and penalty terms, following recent works ~\cite{deng2023syllogistic,pan2019charge}, across four scenarios.
%%%
Charge prediction is assessed as a standard classification task, and we utilize commonly used metrics, i.e., \textit{Accuracy}, \textit{Precision}, \textit{Recall}, and \textit{F1-score}, to evaluate its performance.
%%%
The penalty term prediction is assessed by commonly used \textit{LogD}~\cite{cui2022survey}, which measures the logarithmic difference between the predicted penalty term and the ground truth value.
% It aims to capture subtle variations in their magnitudes.
}

\subsection{Implementation Details}

We fine-tuned benchmark models using a training dataset comprising 4,000 cases for each setting, followed by validation on 500 cases. Subsequently, we evaluated the models using a testing dataset of 500 cases. Additionally, we used prompt templates with and without examples for InternLM2 generation on distinct subtasks, and we applied the same method in a multitask setting.

Following the conclusions in \cite{shui-etal-2023-comprehensive}, we utilized the BM25 \footnote{\url{https://pypi.org/project/rank-bm25/}} retriever to select the most similar case from the test set in each setting, which we then added as an example in our LLM generation. The details of our prompt templates for each setting are provided in Appendix~\ref{sec:prompt-template}.
% experiment result

\section{Outcomes} 
We conduct comprehensive experiments using several benchmark models across four distinct \ac{LJP} scenarios, as shown in Table~\ref{tab:main_results}, and we find that:
\begin{table*}[htb!]
\centering
\footnotesize
\setlength{\tabcolsep}{12pt}
\renewcommand{\arraystretch}{1.2}
\resizebox{\linewidth}{!}{
\begin{tabular}{@{}lc|
>{\centering\arraybackslash}p{1.5cm} 
>{\centering\arraybackslash}p{1.5cm} 
>{\centering\arraybackslash}p{1.5cm} 
>{\centering\arraybackslash}p{1.5cm} 
>{\centering\arraybackslash}p{1.5cm}@{}}
\toprule
\textbf{Metric} & \textbf{Scenario} & \textbf{MT5} & \textbf{BERT} & \textbf{RoBERTa} & \textbf{Lawformer} & \textbf{InternLM2} \\
\midrule
\textbf{Charge Prediction} \\
\midrule
Accuracy (\%) $\uparrow$ & S1 & 75.2 & 78.6 & 81.0 & 81.4 & \textbf{84.6} \\
& S2 & 45.4 & 44.6 & 47.0 & 52.0 & \textbf{80.2} \\
& S3 & 68.8 & 77.8 & 75.2 & 78.0 & \textbf{81.4} \\
& S4 & 30.0 & 29.8 & 30.8 & 34.8 & \textbf{56.2} \\
\midrule
Precision (\%) $\uparrow$ & S1 & 77.7 & 78.6 & 81.0 & 81.4 & \textbf{85.8} \\
& S2 & 77.6 & 67.8 & 71.9 & 73.8 & \textbf{92.1} \\
& S3 & 73.2 & 77.8 & 75.2 & 78.0 & \textbf{81.7} \\
& S4 & 72.7 & 62.8 & 64.1 & 64.1 & \textbf{84.1} \\
\midrule
Recall (\%) $\uparrow$ & S1 & 70.0 & 78.6 & 81.0 & 81.4 & \textbf{84.8} \\
& S2 & 67.2 & 64.9 & 69.1 & 71.0 & \textbf{91.6} \\
& S3 & 68.8 & 77.8 & 75.2 & 78.0 & \textbf{80.4} \\
& S4 & 57.7 & 57.7 & 60.3 & 59.4 & \textbf{77.7} \\
\midrule
F1-Score (\%) $\uparrow$ & S1 & 76.4 & 78.6 & 81.0 & 81.4 & \textbf{85.3} \\
& S2 & 72.0 & 66.3 & 70.5 & 72.4 & \textbf{91.8} \\
& S3 & 70.9 & 77.8 & 75.2 & 78.0 & \textbf{81.0} \\
& S4 & 64.3 & 60.1 & 62.1 & 61.7 & \textbf{80.8} \\
\midrule
\textbf{Penalty Term} \\
\midrule
\multirow{4}{*}{LogD (\%) $\downarrow$} & S1 & 60.7 & 45.8 & 43.3 & \textbf{39.5} & 59.3 \\
& S2 & 62.0 & 51.5 & 49.3 & \textbf{46.4} & 54.1 \\
& S3 & 79.8 & 53.6 & 51.7 & \textbf{48.7} & 61.3 \\
& S4 & 68.3 & 56.8 & 57.1 & 58.5 & \textbf{56.5} \\
\bottomrule
\end{tabular}
}
\caption{Main results of benchmark models across four scenarios (S1–S4), with metrics split into Charge and Penalty Term tasks. Bold font indicates the highest value per row. ``$\uparrow$'' indicates higher is better, ``$\downarrow$'' indicates lower is better.}
\label{tab:main_results}
\end{table*}

\noindent\textbf{Scenario Complexity.} 
We observe that performance consistently declines as scenario complexity increases. 
Specifically, scenario (S4), involving multiple defendants and multiple charges, leads a significant drop in all evaluation metrics across most models, followed by S2 (multiple charges), S3 (multiple defendants), and S1 (single defendant and single charge).
For example, in S4 compared to S1, InternLM2 achieves approximately \todo{4.5\%} lower F1-score and \todo{2.8} higher LogD, while Lawformer demonstrates around \todo{19.7\%} lower F1-score and \todo{19.0} higher LogD.
These results highlight the challenges of modeling complex legal cases, suggesting that methods optimized for simpler scenarios do not generalize well to real-world, multifaceted legal judgments.
% This demonstrates that scenarios involving multiple defendants and charges are still challenging and cannot be treated as simply as the single defendant and/or charge scenarios.

\begin{table*}[htb!]
\centering
\footnotesize
\setlength{\tabcolsep}{12pt}
\renewcommand{\arraystretch}{1.2}
\resizebox{\linewidth}{!}{
\begin{tabular}{@{}lc|cccc@{}}
\toprule
\textbf{Metric} & \textbf{Scenario} &\textbf{Fine-tuning} & \textbf{Multi-task} & \textbf{/wo example} & \textbf{/w example} \\
\midrule
\textbf{Charge Prediction} \\
\midrule
Accuracy (\%) $\uparrow$ & S1 & \textbf{84.6} & 79.0 & 55.4 & 61.8 \\
& S2 & \textbf{80.2} & 79.6 & 37.2 & 58.6 \\
& S3 & \textbf{79.6} & 68.0 & 55.6 & 69.2 \\
& S4 & 56.2 & \textbf{56.4} & 26.0 & 37.2 \\
\midrule
Precision (\%) $\uparrow$ & S1 & \textbf{85.8} & 84.7 & 64.1 & 67.8 \\
& S2 & \textbf{92.1} & 91.0 & 62.1 & 81.9 \\
& S3 & \textbf{81.7} & 77.3 & 70.2 & 77.7 \\
& S4 & \textbf{84.1} & 80.9 & 54.0 & 64.7 \\
\midrule
Recall (\%) $\uparrow$ & S1 & \textbf{84.8} & 80.0 & 55.8 & 62.4 \\
& S2 & \textbf{91.6} & \textbf{91.6} & 73.6 & 82.9 \\
& S3 & \textbf{80.4} & 68.6 & 56.0 & 69.6 \\
& S4 & 77.7 & \textbf{80.6} & 62.1 & 72.6 \\
\midrule
F1-Score (\%) $\uparrow$ & S1 & \textbf{85.3} & 82.3 & 59.7 & 65.0 \\
& S2 & \textbf{91.8} & 91.3 & 67.3 & 82.4 \\
& S3 & \textbf{81.0} & 72.7 & 62.3 & 73.4 \\
& S4 & \textbf{80.8} & 80.7 & 57.8 & 68.4 \\
\midrule
\textbf{Penalty Term Prediction} \\
\midrule
LogD (\%) $\downarrow$ & S1 & 59.3 & \textbf{50.9} & 105.6 & 56.3 \\
& S2 & 54.1 & \textbf{35.8} & 83.8 & 61.7 \\
& S3 & 61.3 & 58.2 & 103.1 & \textbf{56.9} \\
& S4 & \textbf{56.5} & \textbf{53.4} & 84.0 & 70.8 \\
\bottomrule
\end{tabular}
}
\caption{Ablation study of InternLM2 settings across four scenarios (S1–S4). Bold font highlights the best performance for each metric and scenario. ``$\uparrow$'' denotes higher is better, ``$\downarrow$'' denotes lower is better.}
\label{tab:analysis}
\end{table*}

\noindent\textbf{Model Robustness.}
Performance across scenarios also reveals key differences in model robustness.
InternLM2 maintains stable performance despite scenario complexity, whereas less capable models (e.g., RoBERTa, Lawformer) show substantial degradation.
% the impact of scenarios varies significantly depending on specific models. 
% Compared with the top-performing model, InternLM2, the inferior models exhibit larger differences across the scenarios.
For example, Lawformer decreases by 19.7\% in F1-score from (S1) to (S4), while InternLM2 only drops 4.5\%.
This suggests that recent large-scale pre-trained models are better equipped to handle compositional and contextual variance in complex legal documents.
We also conduct an ablation study to further analyze InternLM2 under different training settings in Table~\ref{tab:analysis}.
\noindent\textbf{Training Strategy Analysis.}
We analyze variant training strategies for InternLM2 and find that: 
(i) supervised fine-tuning on each subtasks achieves the best overall performance;
(ii) multi-task learning, while appealing for joint modeling, introduces training difficulties and consistently underperforms fine-tuning.

\noindent\textbf{Prompting Strategies.} We also observe that adding a demonstration example in the prompt improves performance across all scenarios compared to prompting without examples. 
This aligns with findings in in-context learning, highlighting that even a single illustrative example can enhance the model's ability to align with the task, particularly in settings involving few or no training instances.

\section{Discussion}
% \clearpage
% \newpage

\subsection{Limitations}

While our study provides valuable insights, several limitations should be acknowledged. 
First, the dataset, sourced exclusively from Chinese criminal cases, may limit the generalizability of our findings to other legal systems.
% Additionally, our focus on four specific legal scenarios might not encompass all possible situations, and future research should consider a broader range for comprehensive evaluation. 
Second, the complexity of our dataset, especially with multiple defendants and charges, might affect how well models perform. Using a more balanced dataset with different types of cases could help. 
Potential biases in the training data could also affect model fairness, and despite anonymization efforts, data privacy risks remain, necessitating robust techniques and compliance with privacy regulations. 
Third, we notice that model performance varies, with some models struggling in complex scenarios, and the evaluation metrics used may not fully capture the nuances of legal judgments. Improving models through extra fine-tuning or combining different models might reduce this issue. 
Lastly, the black-box nature of LLMs limits their interpretability for understanding how they make decisions, posing challenges for practical use in the legal domain where decision transparency is critical. Developing methods for better transparency and decision justification could address this issue, making the models more usable in practice.
Addressing these limitations is essential for advancing legal judgment prediction and ensuring the ethical and practical deployment of LLMs in the legal field.

% ==============================================================================

\subsection{Ethical Statement}
Throughout this research, we strictly followed ethical guidelines to ensure the responsible use of AI use and protect human data. We closely monitored \acp{LLM} employed to avoid generating harmful or biased content, especially in sensitive areas such as legal judgments.

\noindent\textbf{Data Anonymization and Privacy.}
Data privacy is a top priority in our research. Since part of the data comes from legal judgments and contains sensitive information. To protect the privacy of the individuals involved, we implement strict anonymization procedures for any human data. 
We carefully remove or replace all identifiable information, such as names, addresses, and specific personal details to ensure confidentiality and anonymity.

\noindent\textbf{Ethical Concerns.}
We carefully consider the ethical implications throughout our research and strictly follow the ethical guidelines of the institute. We aim to minimize any potential harm or misuse of the data and individual information. 
Future researchers who wish to use the dataset and findings should also follow these ethical standards, ensuring the data is used responsibly and ethically to advance knowledge in the field.

\noindent\textbf{Use of AI Tools.}
In this work, we utilize AI tools (e.g., ChatGPT and Grammarly), solely for checking grammatical errors.

% \section*{Reproducibility}
% To support the development of research and ensure the reproducibility of our work, we will make the dataset and code available at \url{https://github.com/lololo-xiao/MultiJustice-MPMCP}.

\section{Conclusion}
\jp{
In this paper, we introduce a dataset with four practical scenarios involving various numbers of defendants and charges in Chinese \acl{LJP}. 
We aim to answer whether multiple defendants and charges should be treated separately by comparing experimental results on several benchmark models across different scenarios. 
We find that scenarios involving multiple defendants and/or multiple charges pose great challenges. 
We call for future work in the research community to propose advanced models to facilitate smart legal assistants with real-world cases.
}

\bibliographystyle{splncs04}
\bibliography{references}
%
% \newpage
\appendix
\section{Prompt template}
\label{sec:prompt-template}
Prompt templates for \acp{LLM} to generate outcomes with an example or without an example are shown in Table~\ref{tab:prompt template}.

\begin{CJK}{UTF8}{gbsn} % UTF8 encoding and gbsn (Simplified Chinese) font
\begin{table*}[htb!]
\centering
\small
\resizebox{\linewidth}{!}{
\begin{tabular}{@{}lll@{}}
\toprule
\textbf{Charge Prediction} &  &  \\ \midrule
\begin{tabular}[c]{@{}l@{}}
\textbf{Instruction}: 
请你模拟法官依据下面事实和被告人预测被告的罪名（一/多个）。 
只按照例如的格式回答，不用解释。例如：被告人A其行为构成XX罪。\\ 
Please simulate a judge and predict all the charges (single/multiple) of the defendant based on the following factual description. \\
Respond only in the format provided, without explanation. For example: Defendant A is charged with XX.\\ 
\textbf{Example}: 下面是一个预测被告罪名的例子 Here is an example:\\
被告人 Defendant: B \\
事实 Fact: {[}\textit{Fill based on the retrieval results of BM25}{]} \\
预测 Prediction：被告人B其行为构成XX罪。/ Defendant B is charged with XX. \\
\textbf{Input}:\\ 
被告人 Defendant：{[}\textit{Fill based on the incoming data}{]}\\ 
事实 Fact：{[}\textit{Fill based on the incoming data}{]}
\end{tabular}                                     &  &  \\
\midrule
\textbf{Penalty Term Prediction }&  &  \\
\midrule
\begin{tabular}[c]{@{}l@{}}
\textbf{Instruction}:  
请你模拟法官根据下列事实和被告人预测被告的判决刑期。\\ 
只按照例如的格式回答，不用解释。例如：判处被告 人A有期徒刑X年X个月。 \\ 
Please simulate a judge and predict the penalty term of the defendant based on the following factual description. \\ 
Respond only in the format provided, without explanation. For example: Defendant B is sentenced to X Years X Months. \\ 
\textbf{Example}:
下面是一个预测被告刑期的例子 Here is an example: \\ 
被告人 Defendant: B\\ 
事实 Fact：{[}\textit{Fill based on the retrieval results of BM25}{]} \\ 
预测 Prediction：判处被告人B有期徒刑X月。 / Defendant B is sentenced to X Months. \\ 
\textbf{Input}: \\ 
被告人 Defendant：{[}\textit{Fill based on the incoming data}{]}\\ 
事实 Fact：{[}\textit{Fill based on the incoming data}{]} 
\end{tabular}                  &  &  \\
\midrule
\textbf{Multitask: Charge and Penalty Term Prediction} &  &  \\
\midrule
\begin{tabular}[c]{@{}l@{}}
\textbf{Instruction}: 
请你模拟法官根据下列事实和被告人预测被告的所有罪 名（多个）以及最终判决刑期。\\ 
只按照例如的格式回答， 不用解释。例如：被告人A其行为构成XX罪、XX罪，判处 有期徒刑X年X个月。 \\ 
Please simulate a judge and predict all the charges (single/multiple) and terms penalty of the defendant based on the following factual description. \\ 
Respond only in the format provided, without explanation. For example: Defendant A is charged with XX, and sentenced to X Years X Months. \\ 
\textbf{Example}: 
下面是一个预测被告罪名和刑期的例子 Here is an example: \\ 
被告人 Defendant: B \\ 
事实 Fact: {[}\textit{Fill based on the retrieval results of BM25}{]} \\ 
预测 Prediction：被告人B其行为构成XX罪,被判处有期徒刑X月。 / Defendant A is charged with XX, and sentenced to X Months.\end{tabular} &  &  \\ 
\textbf{Input}: \\ 
被告人 Defendant：{[}\textit{Fill based on the incoming data}{]}\\ 
事实 Fact：{[}\textit{Fill based on the incoming data}{]} \\
\bottomrule
\end{tabular}
}
\caption{Prompt templates used in this paper.}
\label{tab:prompt template}
\end{table*}
\end{CJK}
\end{document}